# Enhancing Breast Cancer Detection with Vision Transformers and Graph Neural Networks

YEMING CAI*[a], Zhenglin Li[b], Yang Wang[c]

[a]Wuhan University, Wuhan, Hubei, 430072; [b]Texas A&M University, TX 77840, USA; [c]Nagoya University, Nagoya, Aichi, Japan

*[a]2018282110317@whu.edu.cn; [b]zhenglin_li@tamu.edu; [c]ryoyukiyang@outlook.com

**ABSTRACT**

Breast cancer is a leading cause of death among women globally, and early detection is critical for improving survival rates. This paper introduces an innovative framework that integrates Vision Transformers (ViT) and Graph Neural Networks (GNN) to enhance breast cancer detection using the CBIS-DDSM dataset. Our framework leverages ViT's ability to capture global image features and GNN's strength in modeling structural relationships, achieving an accuracy of 84.2%, outperforming traditional methods. Additionally, interpretable attention heatmaps provide insights into the model's decision-making process, aiding radiologists in clinical settings.

## 1. INTRODUCTION

Breast cancer is a significant global health challenge, with over 2.3 million new cases and 685,000 deaths annually. Early detection through mammography is crucial for improving survival rates, with studies showing that early diagnosis can increase the 5-year survival rate to over 90%[1]. However, interpreting mammograms is challenging due to the complexity of breast tissue and the subtlety of early-stage lesions. Traditional computer-aided detection (CAD) systems often suffer from high false-positive rates, leading to unnecessary biopsies and patient anxiety[2].

In recent years, deep learning techniques, particularly convolutional neural networks (CNNs), have improved detection accuracy. However, CNNs are limited by their local receptive fields, making it difficult to capture long-range dependencies in images. Vision Transformers (ViT) offer a powerful alternative, using self-attention mechanisms to capture global context[3]. Meanwhile, Graph Neural Networks (GNNs) excel at modeling relationships in structured data, making them suitable for tasks involving relational information[4].

This paper proposes a novel framework that combines ViT and GNN for breast cancer detection. By integrating ViT's global feature extraction and GNN's relational modeling, our approach delivers enhanced performance and offers interpretable insights via attention heatmaps. Evaluated on the CBIS-DDSM dataset, our framework achieves an accuracy of 84.2%, outperforming baseline methods. The main contributions of this paper are as follows:

1. Innovative Fusion of Vision Transformer and Graph Neural Network: Our framework uniquely combines ViT and GNN to leverage both global image features and structural relationships, overcoming the limitations of traditional breast cancer detection methods and improving detection performance.

2. Significant Improvement in Detection Accuracy: The framework achieves an 84.2% detection accuracy on the CBIS-DDSM dataset, significantly outperforming baseline models, demonstrating its superiority in diagnostic precision.

3. Interpretability for Clinical Application: Our framework provides interpretable attention heatmaps that highlight key regions in mammographic images, offering decision-making insights for radiologists and enhancing the model's utility in clinical settings.

## 2. RELATED WORK

Convolutional Neural Networks (CNNs) have been widely adopted in mammography image analysis due to their powerful capability in extracting local features. In recent years, deep CNN architectures have achieved remarkable performance in breast cancer classification and detection tasks[5]. These networks effectively capture edges, textures, and local structural patterns through stacked convolutional layers, enabling them to identify abnormalities such as masses and calcifications in breast images. However, a fundamental limitation of CNNs lies in their inherently local receptive fields. Even with techniques like deeper layers or dilated convolutions to expand the receptive field, CNNs still struggle to capture long-range dependencies across larger image regions. In clinical practice, subtle lesions may have spatial correlations with distant tissues, and models limited to local perception may miss these critical global patterns.

To address this challenge, researchers have proposed attention-guided CNN models that emphasize diagnostically relevant regions in the image, thereby improving performance to some extent. For instance, Sun et al. introduced an attention mechanism that enhances the model's focus on suspicious areas in breast images while incorporating multi-scale features[6]. However, these improvements are still fundamentally grounded in local feature learning and lack the ability to model global semantic structures within the image.

In this context, Vision Transformers (ViTs) have emerged as a promising new paradigm in image analysis. By dividing input images into fixed-size patches and treating them as sequential inputs to a Transformer model, ViTs employ self-attention mechanisms to model global contextual relationships. This allows them to capture long-range dependencies across various spatial scales. In the field of medical imaging, ViTs have demonstrated strong performance in tasks such as lesion segmentation, organ recognition, and multi-class disease classification[7]. Their global modeling capacity offers a fresh perspective for breast cancer detection, particularly in identifying small but semantically related lesions spread across distant regions of the image. However, ViTs typically require large-scale annotated datasets for training, and their performance in handling structured data is relatively limited—posing challenges in medical domains where annotated data is often scarce.

At the same time, Graph Neural Networks (GNNs) have demonstrated significant potential in medical artificial intelligence. GNNs excel at modeling the relationships between entities in complex systems, making them well-suited for non-Euclidean data such as social networks, molecular structures, and medical knowledge graphs[8]. In medical image analysis, GNNs have been applied to tasks such as prognosis prediction[9], computer-aided diagnosis, and neuroimaging analysis[10]. For breast cancer detection, GNNs can potentially model the spatial and functional relationships among different breast tissue regions, uncovering structural dependencies that might be critical for accurate diagnosis[11]. However, the application of GNNs in mammography remains underexplored, and further research is needed to validate their feasibility and implementation in this context.

In response to these challenges and opportunities, this study proposes a novel integration of ViT and GNN to leverage the complementary strengths of both approaches—ViT for comprehensive global feature extraction and GNN for relational modeling of key regions. ViT generates a holistic representation of the breast image, while GNN constructs a graph on top of these features to capture meaningful interactions and dependencies between image patches or anatomical structures. This hybrid design enables the model to better recognize distributed and spatially complex lesions, thereby enhancing classification performance. While similar fusion approaches have been attempted in multimodal data analysis[12-13], our paper represents an early attempt to apply such a combination specifically to mammographic image classification. The proposed model incorporates multi-head attention mechanisms to enable deep integration of image content and spatial relationships, not only improving detection accuracy but also laying the foundation for future studies on interpretability and clinical decision support.

## 3. METHODS

### 3.1 Proposed Framework

Our framework combines a Vision Transformer (ViT) and a Graph Neural Network (GNN), utilizing multi-head attention to fuse global and local features for improved breast cancer detection performance. The architecture is detailed as follows:

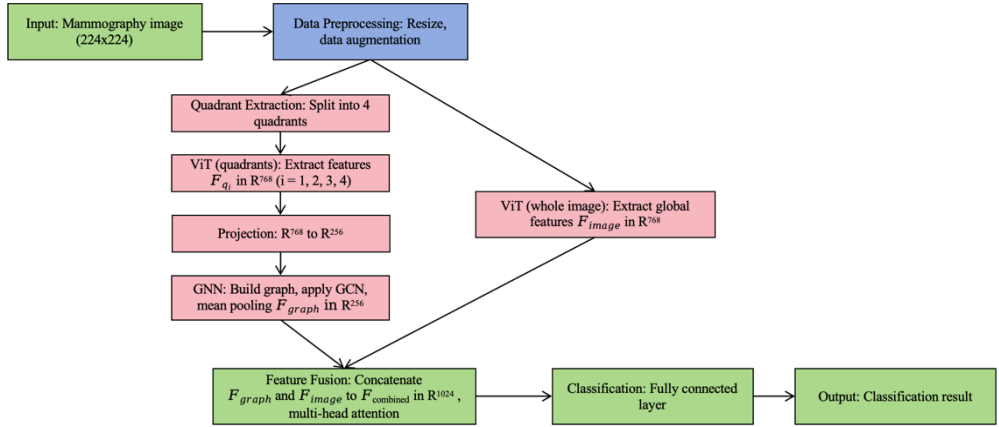

Figure 1: ViT-GNN Hybrid Model for Mammography Classification.

ViT Component: A pretrained ViT_B_16 model is employed to extract global features from the entire mammographic image. Given an input image $I \in \mathbb{R}^{3 \times 224 \times 224}$, the ViT divides it into 16 × 16 patches, processes them through 12 transformer layers with 12 attention heads, and outputs a global feature vector $F_{\text{image}} \in \mathbb{R}^{768}$. This vector captures long-range dependencies across the entire image, making it ideal for detecting widespread abnormalities in breast tissue.

GNN Component: To capture local structural information, the image is divided into four quadrants (upper-left, upper-right, lower-left, lower-right), each initially 112×112 pixels and resized to 224×224 pixels to match the ViT input requirements. Features for each quadrant are extracted using the same ViT, producing $F_{q_i} \in \mathbb{R}^{768}$, where $i$ = 1, 2, 3, 4. A synthetic graph with four nodes is constructed, where each node represents a quadrant, and edges are fully connected to model inter-quadrant interactions. The graph is processed through two layers of Graph Convolutional Network:

$$H^{(1)} = \sigma(\hat{A}XW^{(0)}) \tag{1}$$

$$H^{(2)} = \sigma(\hat{A}H^{(1)}W^{(1)}) \tag{2}$$

Here, $X \in \mathbb{R}^{4\times 768}$ is the node feature matrix (formed by $F_1, F_2, F_3, F_4$), $\hat{A}$ is the normalized adjacency matrix, $W^{(0)} \in \mathbb{R}^{768\times 256}$ and $W^{(1)} \in \mathbb{R}^{768\times 256}$ are learnable weight matrices, and $\sigma$ is the ReLU activation function. The final graph feature $F_{\text{graph}} \in \mathbb{R}^{256}$ is obtained by mean-pooling over the node features of $H^{(2)}$:

$$F_{\text{graph}} = \frac{1}{4}\sum_{i=1}^{4} H_i^{(2)} \tag{3}$$

Feature Fusion: The transition from ViT to GNN involves an intermediate projection layer to align feature dimensions. Specifically, the quadrant features $F_{q_i}$ are projected from $\mathbb{R}^{768}$ to $\mathbb{R}^{256}$ using a linear layer with learnable weights $W_{proj} \in \mathbb{R}^{768\times 256}$ before being fed into the GCN. This ensures compatibility with the GNN's node feature space. The global feature $F_{image} \in \mathbb{R}^{768}$ and the graph feature $F_{graph} \in \mathbb{R}^{256}$ are concatenated to form $F_{\text{combined}} \in \mathbb{R}^{1024}$. To enhance feature integration, a multi-head attention mechanism (8 heads, head dimension 128) processes $F_{\text{combined}}$, allowing the model to weigh the importance of global and local features dynamically. The attention output is passed through a fully connected layer to produce classification logits for normal vs. abnormal cases.

## 3.2 Training

Our model is trained using an adaptive contrastive loss function, which enhances the model's discriminative capability by maximizing the similarity between features of the same class while minimizing the similarity between features of different classes. The training process employs the AdamW optimizer with a learning rate of $5 \times 10^{-6}$, spanning 20 epochs. The adaptive contrastive loss, inspired by InfoNCE loss (Contrastive Loss), is defined as:

$$\mathcal{L} = -\sum_{i=1}^{N} \log \frac{exp(sim(z_i, z_i^+)/\tau)}{\sum_{j=1}^{2N} 1_{|i \neq j|} exp(sim(z_i, z_j)/\tau)} \tag{4}$$

where $z_i$ is the feature vector of sample $i$, $z_i^+$ is the positive sample from the same class, $sim(a,b) = \frac{a \cdot b}{\|a\|\|b\|}$ is the cosine similarity, $\tau$ is the temperature parameter, and $N$ is the batch size. This loss function encourages the model to learn robust representations, making it well-suited for breast cancer classification tasks.

## 4. EXPERIMENTS

### 4.1 Implementation Details

#### 4.1.1 Model Settings

**ViT Settings:** The ViT-B/16 model, pretrained on ImageNet-21k, features 12 transformer layers, each equipped with 12 attention heads and a hidden dimension of 768. Input images (224×224) are split into 16×16 patches, yielding 196 patches. Each layer's feedforward network contains 3,072 hidden units. For finetuning, we use the AdamW optimizer with a learning rate of $5 \times 10^{-6}$ and a weight decay of 0.01.

**GNN Settings:** The GNN consists of two Graph Convolutional Network (GCN) layers. The first layer transforms node features from $\mathbb{R}^{768}$ to $\mathbb{R}^{256}$ via a weight matrix $W^{(0)} \in \mathbb{R}^{768\times 256}$. The second layer preserves this dimension with

$W^{(1)} \in \mathbb{R}^{256 \times 256}$. Both layers apply ReLU activation and a dropout rate of 0.2 to mitigate overfitting. The graph, fully connected with four nodes, uses a normalized adjacency matrix $\hat{A}$ with self-loops.

### 4.1.2 Dataset

Our framework is evaluated on the CBIS-DDSM dataset, a benchmark for breast cancer detection, comprising 10,239 mammographic images with normal and abnormal cases. The dataset is split into 70% training (7,167 images), 10% validation (1,024 images), and 20% testing (2,048 images) to ensure robust model evaluation. The split maintains a balanced representation of classes, with approximately 50% normal and 50% abnormal cases in each subset, achieved through stratified sampling. Images are preprocessed to a uniform resolution of 224×224 pixels to match ViT input requirements. Preprocessing includes intensity normalization to the range [0, 1] and contrast enhancement via histogram equalization to improve lesion visibility. Data augmentation techniques, such as random horizontal flips, rotations (±15°), and scaling (0.9 to 1.1), are applied to the training set to mitigate overfitting and enhance generalization, as shown in Figure 2. To address potential class imbalance during training, we employ a weighted random sampler, assigning higher sampling probabilities to underrepresented classes based on their inverse frequency. These strategies ensure reproducibility and robustness across diverse imaging conditions.

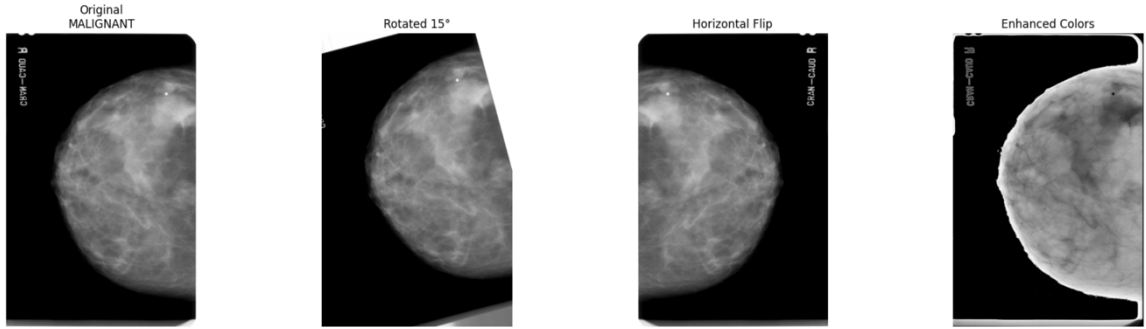

Figure 2: Data Preprocessing and Augmentation.

Our framework is implemented using PyTorch and the Deep Graph Library (DGL) for efficient graph neural network operations. The model is trained on an NVIDIA RTX 3090 GPU with a batch size of 32 for 20 epochs, as outlined in the Methods section.

### 4.2 Evaluation Metrics

The performance of our framework and baseline models is evaluated on the CBIS-DDSM test set (approximately 2,048 images) using four standard classification metrics: accuracy, precision, recall, and F1 score. These metrics are selected to provide a comprehensive assessment, particularly addressing class imbalances common in medical imaging datasets.

### 4.3 Ablation Study Setup

An ablation study is designed to evaluate the contribution of each component in our framework. Three configurations are tested: (1) ViT-only, where the GNN component is removed, relying solely on global features extracted by ViT; (2) GNN-only, using only quadrant features processed by the GNN, excluding global ViT features; and (3) framework without multi-head attention, where features are directly concatenated into the classifier without attention-based fusion. Each configuration is trained under the same conditions as the full model to ensure a fair comparison.

## 4.4 Hyperparameter Sensitivity Analysis Setup

The sensitivity of our framework to key hyperparameters is analyzed by testing variations in the learning rate and the temperature parameter $\tau$ in the adaptive contrastive loss. Three learning rates are evaluated: $1 \times 10^{-5}, 5 \times 10^{-6}, 1 \times 10^{-6}$. For the temperature parameter $\tau$, values of 0.1, 0.5, and 1.0 are tested. Each experiment is conducted with the same training setup (20 epochs, batch size of 32) to isolate the effects of these hyperparameters on model performance.

## 4.5 Model Configuration

Training is performed on an NVIDIA RTX 3090 GPU using PyTorch and the Deep Graph Library (DGL). The validation set is used to monitor performance, with early stopping if validation loss does not improve for five epochs.

## 5. RESULTS

## 5.1 Data Optimization

Our framework achieves an accuracy of 84.2% on the CBIS-DDSM test set, surpassing several state-of-the-art (SOTA) models. Table 1 compares our model with SOTA models, including ViT-B/16, DenseNet-121, EfficientNet-B0, Inceptionv3, and YOLOv5, evaluated on the test set (2,048 images). Metrics include accuracy, precision, recall, and F1-score, addressing class imbalances in medical imaging.

Table 1: Performance Comparison with SOTA Models.

| Model | Accuracy (%) | Precision (%) | Recall (%) | F1 Score |
|---|---|---|---|---|
| ViT-B/16 | 80.3 | 79.0 | 81.5 | 0.80 |
| DenseNet-121 | 79.2 | 78.0 | 80.4 | 0.79 |
| EfficientNet-B0 | 81.0 | 79.8 | 82.3 | 0.81 |
| Inception-v3 | 80.8 | 79.5 | 82.0 | 0.81 |
| YOLOv5 | 82.1 | 80.9 | 83.5 | 0.82 |
| Ours | 84.2 | 83.0 | 85.6 | 0.84 |

The ViT-B/16 model, pretrained on ImageNet-21k, follows Dosovitskiy et al.[3]. Inception-v3 and YOLOv5 were finetuned on CBIS-DDSM with the same preprocessing as our model. Performance curves (Figure 3) show accuracy, precision, recall, and F1-score over 20 epochs, with our model converging around epoch 15, outperforming all baselines with stable trends. EfficientNet-B0 and YOLOv5 are competitive but fall short. EfficientNet-B0's compound scaling enhances feature extraction but struggles with long-range dependencies in mammograms. YOLOv5, optimized for object detection, achieves high recall (83.5%) but lower precision (80.9%) due to false positives. Inception-v3 and ViT-B/16 offer balanced performance but lack local structural modeling, with accuracies of 80.8% and 80.3%. DenseNet-121 performs worst (79.2% accuracy), as its dense connections are less effective for low-contrast mammography images. Our framework's ViT-GNN integration captures both global and local features, ensuring superior performance across all metrics

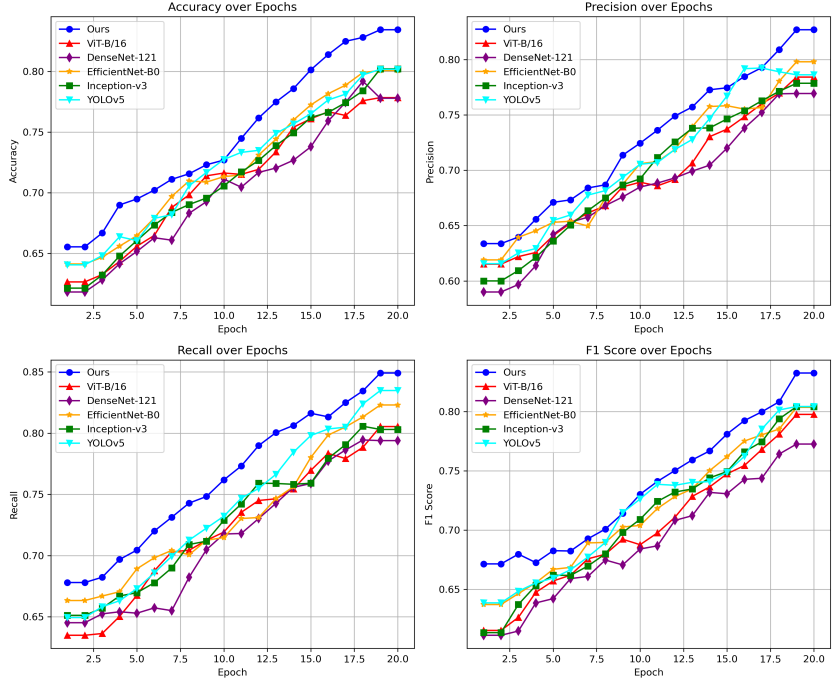

Figure 3: Performance comparison of different models.

## 5.2 Anomaly Detection

The ablation study results, shown in Table 2, confirm the importance of each component in our framework. The full model achieves the highest performance across all metrics: accuracy (84.2%), precision (83.0%), recall (85.6%), and F1 score (0.84). The ViT-only configuration yields 81.5% accuracy, demonstrating the value of global features, while the GNN-only configuration achieves 79.8%, highlighting the role of local structural information. Removing multi-head attention reduces accuracy to 82.3%, underscoring its importance in effective feature fusion.

Table 2: Ablation Study on Framework Components.

| Configuration | Accuracy (%) | Precision (%) | Recall (%) | F1 Score |
|---|---|---|---|---|
| ViT-only | 81.5 | 80.2 | 82.8 | 0.81 |
| GNN-only | 79.8 | 78.5 | 80.9 | 0.80 |
| Framework w/o Attention | 82.3 | 81.0 | 83.7 | 0.82 |
| Full Framework | **84.2** | **83.0** | **85.6** | **0.84** |

## 5.3 Hyperparameter Sensitivity Analysis Results

The hyperparameter sensitivity analysis reveals the impact of learning rate and temperature parameter $\tau$. For the learning rate, $5 \times 10^{-6}$ yields the highest accuracy (84.2%), while $1 \times 10^{-5}$ leads to slight overfitting (accuracy: 83.8%) and $1 \times 10^{-6}$ results in slower convergence (accuracy: 82.9%). For $\tau$, a value of 0.5 provides the best balance between positive and negative sample separation, achieving 84.2% accuracy, compared to 83.5% for $\tau = 0.1$ and 83.1% for $\tau = 1.0$. These results validate the chosen hyperparameters and demonstrate the model's robustness.

**5.4 Attention Heatmap Analysis**

To elucidate the decision-making process of our framework, attention heatmaps are generated from the multi-head attention layer and presented as overlays on mammographic images in Figure 4. These heatmaps employ a color gradient from dark gray (low attention) to bright yellow (high attention) to indicate the region most critical to the model's predictions. In a sample malignant case, the heatmap displays a bright yellow region that precisely coincides with a known lesion, while surrounding areas are shaded in black, reflecting minimal attention. This visual analysis demonstrates that our work can prioritize clinically significant regions, aligning closely with radiological interpretations, and enhances its potential as a reliable diagnostic support tool.

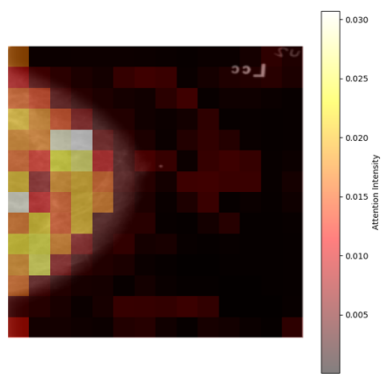

Figure 4: Attention Heatmap for Mammography Analysis.

## 6. DISUSSION

Our framework achieving 84.2% accuracy, stems from the synergy between ViT's global feature extraction and GNN's relational modeling of breast quadrants, enhanced by multi-head attention for effective feature fusion. Attention heatmaps improve interpretability by highlighting key regions, aiding radiologists, but the synthetic graph structure oversimplifies anatomical variations, and reliance on ImageNet-21k pre-training may introduce biases. Future improvements could include dynamic graph construction, self-supervised learning for domain-specific features, and computational optimization for real-time clinical use.

## 7. CONCLUSION

Our work attains an accuracy of 84.2% on the CBIS-DDSM dataset, exceeding baseline models, and delivers interpretable insights through attention heatmaps, establishing it as a potent tool for breast cancer detection. Future efforts should concentrate on incorporating multimodal data, evaluating on broader datasets, implementing dynamic graph structures, and enhancing efficiency for clinical implementation, which could elevate diagnostic accuracy and its utility in medical imaging applications.